\documentclass[11pt]{article}
\usepackage{coling2020}
\usepackage{times}
\usepackage{url}
\usepackage{latexsym}
\usepackage[table]{xcolor}
\usepackage{bm}
\usepackage{graphicx}

\colingfinalcopy 

\title{Combining Deep Learning and String Kernels for the Localization of Swiss German Tweets}

\author{Mihaela G\u{a}man \\
  Department of Computer Science\\
  University of Bucharest\\
  14 Academiei, Bucharest, Romania\\
  {\tt mp.gaman@gmail.com} \\\And
  Radu Tudor Ionescu \\
  Department of Computer Science\\
  Romanian Young Academy\\
  University of Bucharest\\
  14 Academiei, Bucharest, Romania\\
  {\tt raducu.ionescu@gmail.com} \\}

\date{}

\begin{document}
\maketitle

\begin{abstract}
In this work, we introduce the methods proposed by the UnibucKernel team in solving the Social Media Variety Geolocation task featured in the 2020 VarDial Evaluation Campaign. We address only the second subtask, which targets a data set composed of nearly 30 thousand Swiss German Jodels. The dialect identification task is about accurately predicting the latitude and longitude of test samples. We frame the task as a double regression problem, employing a variety of machine learning approaches to predict both latitude and longitude. From simple models for regression, such as Support Vector Regression, to deep neural networks, such as Long Short-Term Memory networks and character-level convolutional neural networks, and, finally, to ensemble models based on meta-learners, such as XGBoost, our interest is focused on approaching the problem from a few different perspectives, in an attempt to minimize the prediction error. With the same goal in mind, we also considered many types of features, from high-level features, such as BERT embeddings, to low-level features, such as characters n-grams, which are known to provide good results in dialect identification. Our empirical results indicate that the handcrafted model based on string kernels outperforms the deep learning approaches. Nevertheless, our best performance is given by the ensemble model that combines both handcrafted and deep learning models.
\end{abstract}

\section{Introduction}
\label{intro}

The organizers of the 2020 VarDial Evaluation Campaign \cite{Gaman-VarDial-2020} proposed a shared task targeted towards the geolocation of short texts, e.g.~tweets, namely the Social Media Variety Geolocation (SMG) task. Typically formulated as a double regression problem, the task is about predicting the location, expressed in latitude and longitude, from where the text received as input was posted on a certain social media platform. Twitter and Jodel are the platforms used for data collection, divided by the language area in three subtasks, namely:
\begin{itemize}
    \item Standard German Jodels (DE-AT) - formed of conversations initiated in Germany and Austria in regional dialectal forms \cite{Hovy-EMNLP-2018}.
    \item Swiss German Jodels (CH) - based on a smaller number of Jodel conversations from Switzerland \cite{Hovy-EMNLP-2018}.
    \item BCMS Tweets - from the area of Bosnia and Herzegovina, Croatia, Montenegro and Serbia where the macro-language used is BCMS, with both similarities and a fair share of variation among the component languages \cite{Ljubesic-COLING-2016}.
\end{itemize}

In this paper, we focus only on the second subtask, SMG-CH, proposing a variety of handcrafted and deep learning models, as well as an ensemble model that combines all our previous models through meta-learning. Our first model is a Support Vector Regression (SVR) classifier \cite{Chang-NC-2002} based on string kernels, which are known to perform well in other dialect identification tasks \cite{Butnaru-VarDial-2018,Ionescu-VarDial-2016,Ionescu-VarDial-2017}. Our second model is a character-level convolutional neural network (CNN) \cite{Zhang-NIPS-2015}, which is also known to provide good results in dialect identification \cite{Butnaru-ACL-2019,Tudoreanu-VarDial-2019}. Due to the high popularity and the outstanding results of Bidirectional Encoder Representations from Transformers (BERT) \cite{Devlin-NAACL-2019} in solving mainstream NLP tasks, we decided to try out a Long Short-Term Memory (LSTM) network \cite{Hochreiter-NC-1997} based on German BERT embeddings as our third model. Lastly, we combine our three models into an ensemble that employs Extreme Gradient Boosting (XGBoost) \cite{Chen-SIGKDD-2016} as meta-learner. We conducted experiments on the development set provided by the organizers, in order to decide which models to choose for our three submissions for the SMG-CH subtask. Our results indicate that the ensemble model attains the best results. Perhaps surprisingly, our shallow approach based on string kernels outperforms both deep learning models. Our observations are consistent across the development and the test sets provided by the organizers.





The rest of this paper is organized as follows. We present related work on dialect identification and geolocation of short texts in Section~\ref{sec_related}. Our approaches are described in more detail in Section~\ref{sec_method}. We present the experiments and empirical results in Section~\ref{sec_experiments}. Finally, our conclusions are drawn in Section~\ref{sec_conclusion}.

\section{Related Work}
\label{sec_related}

One of the initial works on text-based geotagging \cite{Ding-2000} aims at automatically finding the geographic scope of web pages, in a classification setup relying on named location entities such as cities and states. The authors used gazetteers as the source of the location mappings, proposing a rather heuristic approach. Gazetteers, constitute a tool used in one of the three general approaches taken so far in text-based geolocation, this tool being adopted in a number of works \cite{Lieberman-ICDE-2010,Quercini-SIGSPATIAL-2010,Cheng-ACM-2010}. In this line of research, some researchers employed rule-based methods \cite{Bilhaut-NAACL-2003}, while others plugged named entity recognition into various machine learning techniques \cite{Gelernter-2011,Qin-SIGSPATIAL-2010}. The main disadvantage of these methods is that they rely on the existence of specific mentions of locations in text, rather than inferring them in a not so straightforward manner. These direct mentions of places do not represent a safe assumption, especially when it comes to social media platforms such as Twitter, which is used as the data source in some of these studies \cite{Cheng-ACM-2010}. 
The other two main categories of approaches for text-based geolocation rely on either unsupervised learning \cite{Ahmed-WWW-2013,Hong-WWW-2012,Eisenstein-EMNLP-2010} or supervised classification \cite{Wing-HLT-2011,Kinsella-2011}. The unsupervised methods can be described in large part as clustering techniques based on topic models.

There are some studies on user geolocation in social media, that look at this task from a supervised learning perspective \cite{Rout-ACM-2013} and can be included in the second set of approaches for geotagging. However, in such works, other details (e.g.~social ties) in the users profile have been considered rather than their written content. Although these works cover geolocation prediction in social media, they do not use text as input. Our current interest in studying language variation for the geolocation of users in social media has been covered in the literature in a series of works \cite{Rahimi-Arxiv-2017,Han-AIR-2014,Doyle-EACL-2014,Roller-EMNLP-2012,Eisenstein-EMNLP-2010}, employing various machine learning techniques, that range from probabilistic graphical models \cite{Eisenstein-EMNLP-2010} and adaptive grid search \cite{Roller-EMNLP-2012} to Bayesian methods \cite{Doyle-EACL-2014} and neural networks \cite{Rahimi-Arxiv-2017}. 

The related work to date covers a wide range of languages and dialects, including Dutch \cite{Wieling-PLoS-2011}, British \cite{Szmrecsanyi-IJHAC-2008}, American \cite{Huang-CEUS-2016,Eisenstein-EMNLP-2010} and even African American Vernacular English \cite{Jones-AS-2015}. Most related to our work is the study of \newcite{Hovy-EMNLP-2018}, which targets the German language and its variations and, in addition to the previously mentioned endeavours, performs a quantitative analysis against a dialect map. Moreover, \newcite{Hovy-EMNLP-2018} collected 16.8 million online posts from the German-speaking area with the aim of learning document representations of cities. Among these posts, some were from the German speaking side of Switzerland, being part of the SMG shared task, more specifically the SMG-CH subtask that we are addressing. The authors aimed at capturing enough regional variations in the written language, serving as input in automatically distinguishing the geographical region of speakers. The focus was on larger regions covering a given dialect, the proposed approach being based on clustering. Given the shared task formulation, we take a different approach and use the provided data in a double regression setup, addressing the problem both from a shallow perspective and a deep learning perspective, respectively.

\section{Methods}
\label{sec_method}

\subsection{$\bm{\nu}$-Support Vector Regression based on String Kernels.}
\label{subsect:nusvr}
\noindent
\textbf{String Kernels.} \newcite{Lodhi-NIPS-2001} introduced string kernels as a means of comparing two documents, based on the inner product generated by all substrings of length $n$, typically known as character n-grams. 
Since then, string kernels have found many applications, from sentiment analysis \cite{Gimenez-EACL-2017,Ionescu-EMNLP-2018,Popescu-KES-2017}, automated essay scoring \cite{Cozma-ACL-2018} and sentence selection \cite{Masala-KES-2017} to native language identification \cite{Ionescu-EMNLP-2014,Ionescu-COLI-2016,Ionescu-BEA-2017,Popescu-BEA8-2013} and dialect identification \cite{Butnaru-VarDial-2018,Butnaru-ACL-2019,Ionescu-VarDial-2017}.

In this work, we employ string kernels as described in \cite{Butnaru-ACL-2019}, specifically using the efficient algorithm for building string kernels of \newcite{Popescu-KES-2017}. We note that the number of character n-grams is usually much higher than the number of samples, so representing the text samples as feature vectors may require a lot of space. String kernels provide an efficient way to avoid storing and using the feature vectors (primal form), by representing the data though a kernel matrix (dual form). Each cell in the kernel matrix represents the similarity between some text samples $x_i$ and $x_j$. In our experiments, we use the presence bits string kernel \cite{Popescu-BEA8-2013} as the similarity function. For two strings $x_i$ and $x_j$ over a set of characters $S$, the presence bits string kernel is defined as follows:
\begin{equation}\label{eq_str_kernel_presence}
k^{0/1}(x_i, x_j)=\sum\limits_{g \in S^n} \mbox{\#}(x_i, g) \cdot \mbox{\#}(x_j, g),
\end{equation}
where $n$ is the length of n-grams and $\mbox{\#}(x,g)$ is a function that returns 1 when the number of occurrences of n-gram $g$ in $x$ is greater than 1, and 0 otherwise.

\noindent
\textbf{$\bm{\nu}$-Support Vector Regression.} Support Vector Machines (SVM) \cite{Cortes-ML-1995} represent a popular method initially designed for binary classification, which was subsequently repurposed for regression, under the SVR \cite{Drucker-NIPS-1997} acronym (i.e.~Support Vector Regression). Similar to SVM, SVR uses the notion of support vectors and margin in order to find an optimal estimator.  
In the original $\epsilon$-SVR formulation \cite{Drucker-NIPS-1997}, there is an $\epsilon$-insensitive region, i.e. $\epsilon$ tube, defined in the optimization function. The goal is to find the flattest tube containing most of the training samples, while also minimizing the prediction error and model complexity. Different from linear regression, $\epsilon$-SVR fits the error within the maximum margin $\epsilon$, instead of minimizing the error directly \cite{Smola-SC-2004}. In our experiments, we employ an equivalent SVR formulation known as $\nu$-SVR \cite{Chang-NC-2002}, where $\nu$ is the configurable proportion of support vectors to keep with respect to the number of samples in the data set. In $\nu$-SVR, the margin $\epsilon$ is automatically estimated to its optimal value. Using $\nu$-SVR, the optimal solution can converge to a small model, with only a few support vectors. This is especially useful in our use case, as the data set provided in the SMG-CH subtask does not contain too many samples. Another reason to employ $\nu$-SVR in our regression task is that it was found to surpass other regression methods in complex word identification \cite{Butnaru-BEA-2018}.

\subsection{Character-Level Convolutional Neural Network}
\label{subsect:charcnn}

\noindent
\textbf{Character Embeddings.} From the pioneering works in language modelling at the character level \cite{Gasthaus-DCC-2010,Wood-ICML-2009} to date \cite{Georgescu-ArXiv-2020}, a broad range of neural architectures rely on characters as features. Among these, we can mention Recurrent Neural Networks (RNNs) \cite{Sutskever-ICML-2011}, LSTM networks \cite{Ballesteros-EMNLP-2015}, CNNs \cite{Kim-AAAI-2016,Zhang-NIPS-2015} and transformer models \cite{Al-Rfou-AAAI-2019}. 
Characters are the base units in building words that exist in the vocabulary of most languages. 
Knowledge of words, semantic structure or syntax is not required when working with characters. Robustness to spelling errors and words that are outside the vocabulary \cite{Ballesteros-EMNLP-2015} constitute other advantages explaining the growing interest in using characters as features. In our paper, we employ a convolutional neural network working at the character level \cite{Zhang-NIPS-2015}. The employed CNN is equipped with a character embedding layer, automatically learning a 2D representation of text formed of character embedding vectors, that is further processed by the convolutional layers. 

\noindent
\textbf{Convolutional Neural Networks.} Inspired by the visual cortex of mammals \cite{Fukushima-BC-1980}, CNNs have been extensively used in image classification \cite{LeCun-NC-1989,LeCun-CVPR-2004,Krizhevsky-NIPS-2012}, subsequently being adapted for various NLP tasks \cite{Kim-EMNLP-2014,Zhang-NIPS-2015}. CNNs are composed of convolutional blocks, consisting of convolutions and pooling operations, usually followed by a sequence of dense layers and ending with an output layer, with the number of neurons equal to the number of values that we are interested in predicting. In the experiments, we employ a character-level CNN \cite{Zhang-NIPS-2015} with squeeze-and-excitation (SE) blocks, introduced by \newcite{Butnaru-ACL-2019}. Since this method has been previously applied in Romanian dialect identification with good results \cite{Butnaru-ACL-2019,Tudoreanu-VarDial-2019}, we consider it a good candidate for our text geolocation task. We therefore change the original architecture by replacing the Softmax classification layer with a regression layer formed of two units, one predicting the latitude and one predicting the longitude, respectively. We train our character-level CNN towards minimizing the mean squared error (MSE) loss function with respect to the ground-truth latitude and longitude.

\subsection{Long Short-Term Memory Networks based on BERT Embeddings}
\label{subsect:lstm}

\noindent
\textbf{BERT Embeddings.}
Transformers \cite{Vaswani-NIPS-2017} represent a very important advance in Natural Language Processing, with many benefits over the traditional sequential neural architectures. Based on an encoder-decoder architecture with attention, transformers proved to be better at modelling long-term dependencies in sequences, while being effectively trained as the sequential dependency of previous tokens is removed. Unlike other contemporary attempts at using transformers in language modelling \cite{Radford-Arxiv-2018}, BERT \cite{Devlin-NAACL-2019} incorporates context from both directions in the process of building deep language representations, in a self-supervised fashion. The masked language modeling technique enables BERT to pre-train these deep bidirectional representations, that can be further fine-tuned and adapted for a variety of tasks, without significant architectural updates. We also make use of this property in the current work, employing a TensorFlow version of a German BERT model\footnote{\url{https://github.com/deepset-ai/FARM}}. The model has been trained on the latest German Wikipedia dump, the OpenLegalData dump and a collection of news articles, summing up to a total of $12$ GB of text files. We add the pre-trained German BERT model to be fine-tuned in an end-to-end fashion along with our LSTM architecture for geolocation of Swiss German short texts. 

\noindent
\textbf{Long Short-Term Memory Networks.} RNNs \cite{Werbos-NN-1988} operate at the sequence level, attaining state-of-the-art performance on various problems involving time series \cite{Weiss-ACL-2018}. Major drawbacks in regular RNNs are the phenomenons of exploding and vanishing gradients, which can be caused by an increase in the length of the input sequence \cite{Hochreiter-FGDRNN-2001}. LSTM networks \cite{Hochreiter-NC-1997} represent a flavour of RNN, designed to overcome the aforementioned challenges faced when working with RNNs. An LSTM unit has a more complex structure, including a memory cell to remember dependencies in the input and three gates acting as regulators: input, output and, in more recent versions, forget gates, which enable the cell to reset its state \cite{Gers-NC-2000}. The LSTM architecture used in this work is inspired by the one described in \cite{Onose-VarDial-2019}, which has been successfully employed in Romanian dialect identification. We train our LSTM model using the mean squared logarithmic error as loss function. We opted for the aforementioned loss in favor of the mean squared error, as the latter loss function did not produce optimal results for our LSTM.

\subsection{Ensemble Learning}
\label{subsect:xgboost}

\noindent
\textbf{XGBoost.} Gradient tree boosting \cite{Friedman-2001} is based on training a tree ensemble model in an additive fashion. This technique has been successfully used in classification \cite{Li-UAI-2010} and ranking \cite{Burges-Learning-2010} problems, obtaining notable results in reputed competitions such as the Netflix Challenge \cite{Bennett-KDD-2007}. Furthermore, gradient tree boosting is the ensemble method of choice in real-world pipelines running in production \cite{He-DMOA-2014}. XGBoost \cite{Chen-SIGKDD-2016} is a tree boosting model targeted at solving large-scale tasks with limited computational resources. This approach aims at parallelizing tree learning while also trying to handle various sparsity patterns. Overfitting is addressed through shrinkage and column subsampling. Shrinkage acts as a learning rate, reducing the influence of each individual tree. Column subsampling is borrowed from Random Forests \cite{Breiman-ML-2001}, bearing the advantage of speeding up the computations. In the experiments, we employ XGBoost as a meta-learner over the individual predictions of each of the models described above. We opted for XGBoost in detriment of average voting and a $\nu$-SVR meta-learner, both providing comparatively lower performance levels in a set of preliminary ensemble experiments.

\section{Experiments}
\label{sec_experiments}

\subsection{Data Set}
\label{sect:data}

The data set for the SMG-CH subtask contains a training set of 22,600 samples, with one sample per line, each formed of a piece of text and a pair of coordinates representing the position on Earth, i.e.~latitude and longitude. The development set is composed of 3,086 samples, provided in the same format. The test set consists in 3,097 samples without coordinates. We note that the centroid computed on the training data has a latitude of 47.26 degrees and a longitude of 8.33 degrees, confirming that the average location is on the territory of Switzerland.

\subsection{Parameter Tuning}
\label{sect:tuning}

\noindent
\textbf{$\bm{\nu}$-SVR based on string kernels.} In the experiments, we use $\nu$-SVR with a pre-computed string kernel, employing the efficient algorithm proposed in \cite{Popescu-KES-2017}. In a set of preliminary experiments, we employed various blended spectrum string kernels based on various n-gram ranges that include {n-grams} from 3 to 7 characters long. The best performance in terms of both mean absolute error (MAE) and mean squared error (MSE) were attained by a string kernel based on the blended spectrum of 3 to 5 character n-grams. These results are consistent with those reported by \newcite{Ionescu-VarDial-2017}, suggesting that the 3-5 n-gram range is optimal for German dialect identification. The resulting kernel matrix is used as input in a double regression setup, with a $\nu$-SVR model for predicting the latitude (in degrees), and another $\nu$-SVR model for predicting the longitude (in degrees), respectively. We tried out values ranging from $10^{-4}$ to $10^{4}$ for the regularization penalty $C$, during the hyperparameter tuning phase. Similarly, for the proportion of support vectors $\nu$, we considered $10$ values covering the interval $(0, 1]$ with a step of $0.1$. For both regression models, the best value for the parameter $C$ is $10$. As for the parameter $\nu$, the default value of $0.5$ seems to yield the best results.  

\noindent
\textbf{Character-level CNN.} Except for the last layer, the architecture used in our experiments is identical to the architecture employed by \newcite{Butnaru-ACL-2019} for Romanian dialect identification. An input of maximum $5000$ characters (zero-padding is used as necessary) is expected into the network, with the characters initially encoded with their position in the vocabulary. For each character, a vectorial representation of $128$ elements is learned in the embedding layer. Three convolutional blocks follow, each being composed of a convolutional layer with $128$ one-dimensional filters of size $7$ for the first two blocks, and of size $3$ for the last block, resptectively. Each convolutional block also performs downsampling through max-pooling operations with a filter of size $3$. Squeeze-and-excitation (SE) attention modules are integrated after each pooling layer. The outputs are then flattened and given as input into the regression layer, containing two neurons, one for predicting the latitude and the other for predicting the longitude. Adam \cite{Kingma-ICLR-2014} is used as the optimization algorithm, in an attempt to minimize the MSE loss. We trained our model on mini-batches of $128$ samples for $100$ epochs with early stopping, using a learning rate of $5\cdot 10^{-4}$. The network converged in $60$ epochs, after observing no improvements for the last $7$ epochs. 

\noindent
\textbf{LSTM based on BERT embeddings}. In conjunction with the LSTM, we fine-tuned a BERT model that is pre-trained on a German corpus, as detailed in Section~\ref{subsect:nusvr}. Thus, we input the data into a BERT layer, initialized with the corresponding pre-trained parameters. We set the maximum sequence length to $310$, which is around the mean sequence length in the SMG-CH data set. The BERT layer is followed by two LSTM layers of size $128$ each, both having \emph{tanh} activation. We use dropout for regularization, randomly removing $20\%$ of the neurons. We tried out various optimization algorithms such as Adam, RMSProp and stochastic gradient descent (SGD) with momentum. We obtained the best convergence using SGD with a momentum rate of $0.9$ and a learning rate of $\alpha=10^{-1}$. The training automatically ended after $18$ epochs because of early stopping, as there were no more improvements registered for the loss value. 

\noindent
\textbf{Extreme Gradient Boosting}. We employed XGBoost as a meta-learner, training it over the predictions of all the other models. In our case, XGBoost provided the best results with the number of estimators set to $1000$, a maximum depth of $10^{-4}$ and the learning rate $\alpha = 10^{-2}$.

\subsection{Preliminary Results}
\label{sect:results}

In the development phase, as there was no metric specified in the description of the SMG task, we treated it as any other regression problem and evaluated the predictions in terms of both MAE and MSE. In Table~\ref{tab_results_mae_mse}, we present the results obtained by our four models on the development set. As the organizers released the ground-truth labels for the test set after the competition, we also include the MAE and the MSE on the test set, for reference.

\begin{table}[!h]
\begin{center}
\caption{Results in terms of Mean Absolute Error (MAE) and Mean Squared Error (MSE) obtained on the development set and on the test set by the proposed handcrafted, deep and ensemble algorithms. The reported MAE and MSE values represent the average value computed over the latitude and the longitude, both being expressed in degrees.}
\vspace{0.2cm}
\label{tab_results_mae_mse}
\begin{tabular}{ccccc}
\hline
Method                                & \multicolumn{2}{c}{MAE}        &  \multicolumn{2}{c}{MSE} \\

\cline{2-5}
                              &  Development    &  Test        &  Development    &  Test   \\

\hline
\rowcolor{gray!20!white}
$\nu$-SVR + string kernels             &  0.2306        &  0.2289       &  0.1066         &  0.1049 \\ 
character-level CNN                 &  0.2937        &  0.3123       &  0.1552         &  0.1633 \\
\rowcolor{gray!20!white}
LSTM      + BERT embeddings                      &  0.3594        &  0.3618       &  0.2226         &  0.2259 \\
XGBoost ensemble                        &  0.2234        &  0.2207       &  0.1043         &  0.1017 \\
\hline
\end{tabular}
\end{center}
\end{table}

Considering the results presented in Table~\ref{tab_results_mae_mse}, it is clear that the algorithm that achieves the best MAE and MSE values is the ensemble based on the XGBoost meta-learner. This does not come as a surprise, as the ensemble combines the predictions of three individual models, each being based on a different type of features and a different learning model. While these aspects are complementary in theory, the results indicate that this is also the case in practice. 

Additionally, we note that the performance achieved by $\nu$-SVR comes close to the one attained by the ensemble, leaving behind the two neural models based on characters and fine-tuned BERT embeddings. This confirms the efficiency of string kernels over deep learning approaches observed in related works~\cite{Butnaru-ACL-2019,Gaman-ArXiv-2020}, which seems to be independent of the task to be solved.

Another comment regarding the results outlined in Table~\ref{tab_results_mae_mse} is that the errors on the development set do not fall far from the ones obtained on the test set. However, the ensemble model as well as the $\nu$-SVR based on string kernels obtain slightly lower errors at test time compared to the errors reported on the development set. The opposite seems to happen with the deep models, as both the LSTM based on BERT embeddings and the character-level CNN yield slightly higher errors on the test data than on the development data.

Every participant was allowed to make three submissions to compete against other shared task participants. Based on the results reported on the development set, we have decided to choose the character-level CNN, the $\nu$-SVR based on string kernels and the XGBoost ensemble as candidates for the SMG-CH challenge. We excluded the LSTM based on BERT embeddings, since it attains the highest errors among the considered models.

\subsection{Final Results}

\begin{table}[!h]
\begin{center}
\caption{The final results of our team (UnibucKernel) obtained in the SMG-CH subtask, with the metrics picked by the organizers, oriented on clustering by city and on distances expressed in kilometers.}
\label{tab_results_final}
\vspace{0.2cm}
\begin{tabular}{cccccc}
\hline
Method    & Submission     & Median distance  & Mean distance & Clustering Accuracy  \\
\hline
\rowcolor{gray!20!white}
character-level CNN     & \#1  & 40.23             & 42.87         & 29.79\% \\
$\nu$-SVR + string kernels   & \#2 & 26.78             & 31.49         & 51.13\% \\
\rowcolor{gray!20!white}
XGBoost   ensemble      & \#3       & 25.57             & 30.52         & 53.88\% \\ 
\hline
\end{tabular}
\end{center}
\end{table}

Table \ref{tab_results_final} shows our final results obtained on the test set, considering the metrics chosen by the organizers, which are oriented on distances (in kilometers) and on clustering accuracy. Considering the official metrics, our best submission placed us in the top six participants.

We observe that our best performing algorithm, namely the XGBoost ensemble combining the predictions of both deep and shallow methods based on various types of features, achieves a median distance of $25.57$ \emph{km}, a mean distance of $30.52$ \emph{km} and a clustering accuracy of $53.88\%$. Consistent with our findings on the development set, the $\nu$-SVR based on string kernels does not fall far behind the XGBoost ensemble, obtaining a mean distance and median distance that is about one kilometer higher and a clustering accuracy that is nearly $2.75\%$ lower. The deep character-level CNN seems significantly worse, with around $15$ \emph{km} and more than $10$ \emph{km} higher errors in terms of the median and the mean distances as compared to the other two submitted models, and around $21\%$ lower clustering accuracy. In our opinion, these results stand proof that neural networks might not be the holy grail in every possible situation.

\begin{figure}[!t]
\centering
\includegraphics[width=1.0\linewidth]{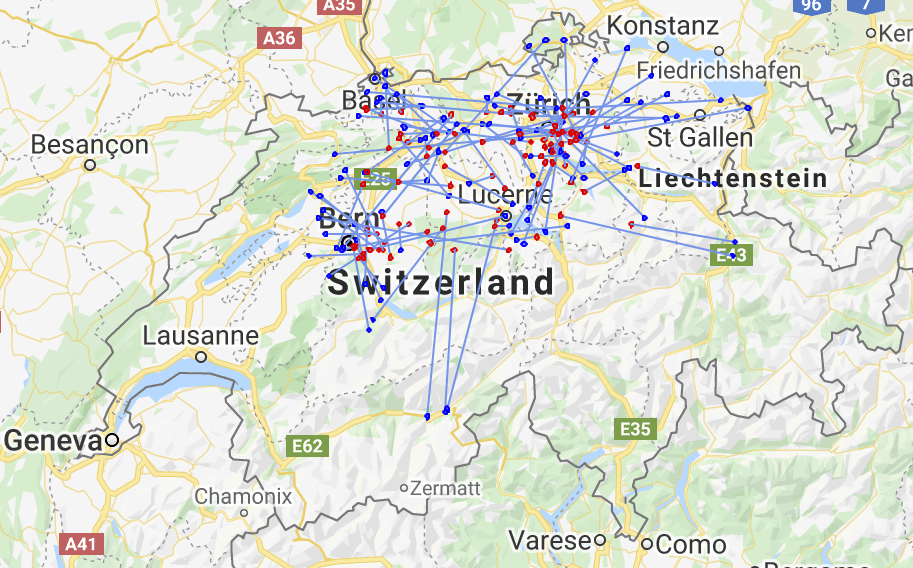}
\caption{Distances between ground-truth locations (blue) and predicted locations (red) for a subset of $100$ Swiss German Jodels randomly selected from the official test set. Best viewed in color.}
\label{fig_map_random_preds}
\end{figure}

We believe that the proposed methods attain decent results, given the challenging nature of the problem at hand. However, it is clear to us that they could benefit from some improvements, considering the values obtained in the final evaluation phase. One important step in this direction is to visualize the errors, at scale, on the map of Switzerland, for a better understanding of the patterns that our best performing algorithm drew with its predictions. Thus, we overlap the predicted and the ground-truth locations on the map of Switzerland, illustrating the result in Figure~\ref{fig_map_random_preds}. The points depicted on the map are described by their 2D coordinates, latitude and longitude, as in the data set provided for the task. The ground-truth locations are colored in blue, while the predictions are illustrated in red. For each pair of ground-truth and predicted location, there is a line connecting the two points, giving us a better idea regarding the errors made by the XGBoost ensemble, in terms of distance. For the visualization presented in Figure~\ref{fig_map_random_preds}, we have randomly selected a subset of $100$ points from the test set, with non-overlapping ground-truth locations. We hereby notice that including all the data points from the test set would generate a visualization that is too cluttered and hard to understand. Hence, we opted for a smaller number of points for a better visualization experience. Considering the annotated map illustrated in Figure~\ref{fig_map_random_preds}, we observe that our predictions tend to be clustered around the main cities in the German-speaking side of Switzerland, such as Z\"{u}rich, Bern, Lucerne and Basel. This bias towards the mentioned cities might be induced by the data samples from the training set, likely not having a well-distributed variance in terms of the locations or the texts used in the learning process. We also observe that the ground-truth locations exhibit a higher variance than the predicted locations. One possible solution would to manually adjust the variance of the predicted location to match the variance of the actual locations.

\section{Conclusion}
\label{sec_conclusion}

In the current work, we tackled the SMG-CH shared subtask of the 2020 VarDial Evaluation Campaign. We addressed this challenge from a shallow perspective, with handcrafted models such as a $\nu$-SVR based on string kernels, as well as from a deep learning perspective, with neural models such as an LSTM based on BERT embeddings and a character-level CNN, respectively. Additionally, we combined the proposed models into an ensemble, employing the XGBoost meta-learner. We obtained our best results with the XGBoost ensemble, which benefits from complementary information from the handcrafted and deep models. We therefore brought one more proof regarding the effectiveness of ensemble learning in general, and of XGBoost, in particular. Another important conclusion is that our shallow model based on string kernels outperforms the two deep neural networks. We consider this as yet another indicator of the high discriminative power that string kernels can bring to a fairly standard learning model, i.e.~the $\nu$-SVR. 

In future work, we aim to explore ways to improve our performance with respect to the metrics proposed by the shared task organizers. Currently, it seems that training the models to simply minimize the MSE or the MAE values is not effective, as our best model was significantly outperformed by the model proposed by the shared task organizers themselves.

\section*{Acknowledgements}

This work was supported by a grant of the Romanian Ministry of Education and Research, CNCS - UEFISCDI, project number PN-III-P1-1.1-TE-2019-0235, within PNCDI III. This article has also benefited from the support of the Romanian Young Academy, which is funded by Stiftung Mercator and the Alexander von Humboldt Foundation for the period 2020-2022.

\bibliographystyle{coling}
\bibliography{coling2020}

\begin{thebibliography}{}

\bibitem[\protect\citename{Ahmed \bgroup et al.\egroup }2013]{Ahmed-WWW-2013}
Amr Ahmed, Liangjie Hong, and Alex~J. Smola.
\newblock 2013.
\newblock Hierarchical geographical modeling of user locations from social
  media posts.
\newblock In {\em Proceedings of WWW}, pages 25--36.

\bibitem[\protect\citename{Al-Rfou \bgroup et al.\egroup
  }2019]{Al-Rfou-AAAI-2019}
Rami Al-Rfou, Dokook Choe, Noah Constant, Mandy Guo, and Llion Jones.
\newblock 2019.
\newblock {Character-Level Language Modeling with Deeper Self-Attention}.
\newblock In {\em Proceedings of AAAI}, pages 3159--3166.

\bibitem[\protect\citename{Ballesteros \bgroup et al.\egroup
  }2015]{Ballesteros-EMNLP-2015}
Miguel Ballesteros, Chris Dyer, and Noah~A. Smith.
\newblock 2015.
\newblock {Improved Transition-Based Parsing by Modeling Characters instead of
  Words with LSTMs}.
\newblock In {\em Proceedings of EMNLP 2015}, pages 349--59.

\bibitem[\protect\citename{Bennett \bgroup et al.\egroup
  }2007]{Bennett-KDD-2007}
James Bennett, Stan Lanning, et~al.
\newblock 2007.
\newblock {The Netflix Prize}.
\newblock In {\em Proceedings of KDD}, volume 2007, page~35.

\bibitem[\protect\citename{Bilhaut \bgroup et al.\egroup
  }2003]{Bilhaut-NAACL-2003}
Fr{\'e}d{\'e}rik Bilhaut, Thierry Charnois, Patrice Enjalbert, and Yann Mathet.
\newblock 2003.
\newblock Geographic reference analysis for geographic document querying.
\newblock In {\em Proceedings of HLT-NAACL-GEOREF}, pages 55--62.

\bibitem[\protect\citename{Breiman}2001]{Breiman-ML-2001}
Leo Breiman.
\newblock 2001.
\newblock Random forests.
\newblock {\em Machine learning}, 45(1):5--32.

\bibitem[\protect\citename{Burges}2010]{Burges-Learning-2010}
Christopher~J.C. Burges.
\newblock 2010.
\newblock {From RankNet to LambdaRank to LambdaMART: An Overview}.
\newblock {\em Learning}, 11(23-581):81.

\bibitem[\protect\citename{Butnaru and Ionescu}2018a]{Butnaru-BEA-2018}
Andrei Butnaru and Radu~Tudor Ionescu.
\newblock 2018a.
\newblock {UnibucKernel: A kernel-based learning method for complex word
  identification}.
\newblock In {\em Proceedings of BEA-13}, pages 175--183.

\bibitem[\protect\citename{Butnaru and Ionescu}2018b]{Butnaru-VarDial-2018}
Andrei~M. Butnaru and Radu~Tudor Ionescu.
\newblock 2018b.
\newblock {UnibucKernel Reloaded: First Place in Arabic Dialect Identification
  for the Second Year in a Row}.
\newblock In {\em Proceedings of VarDial}, pages 77--87.

\bibitem[\protect\citename{Butnaru and Ionescu}2019]{Butnaru-ACL-2019}
Andrei~M. Butnaru and Radu~Tudor Ionescu.
\newblock 2019.
\newblock {MOROCO: The Moldavian and Romanian Dialectal Corpus}.
\newblock In {\em Proceedings of ACL}, pages 688--698.

\bibitem[\protect\citename{Chang and Lin}2002]{Chang-NC-2002}
Chih-Chung Chang and Chih-Jen Lin.
\newblock 2002.
\newblock {Training {$\nu$}-Support Vector Regression: Theory and Algorithms}.
\newblock {\em Neural Computation}, 14:1959--1977.

\bibitem[\protect\citename{Chen and Guestrin}2016]{Chen-SIGKDD-2016}
Tianqi Chen and Carlos Guestrin.
\newblock 2016.
\newblock {XGBoost: A scalable tree boosting system}.
\newblock In {\em Proceedings of KDD}, pages 785--794.

\bibitem[\protect\citename{Cheng \bgroup et al.\egroup }2010]{Cheng-ACM-2010}
Zhiyuan Cheng, James Caverlee, and Kyumin Lee.
\newblock 2010.
\newblock You are where you tweet: a content-based approach to geo-locating
  twitter users.
\newblock In {\em Proceedings of CIKM}, pages 759--768.

\bibitem[\protect\citename{Cortes and Vapnik}1995]{Cortes-ML-1995}
Corinna Cortes and Vladimir Vapnik.
\newblock 1995.
\newblock Support-vector networks.
\newblock {\em Machine Learning}, 20(3):273--297.

\bibitem[\protect\citename{Cozma \bgroup et al.\egroup }2018]{Cozma-ACL-2018}
M\u{a}d\u{a}lina Cozma, Andrei Butnaru, and Radu~Tudor Ionescu.
\newblock 2018.
\newblock Automated essay scoring with string kernels and word embeddings.
\newblock In {\em Proceedings of ACL}, pages 503--509.

\bibitem[\protect\citename{Devlin \bgroup et al.\egroup
  }2019]{Devlin-NAACL-2019}
Jacob Devlin, Ming-Wei Chang, Kenton Lee, and Kristina Toutanova.
\newblock 2019.
\newblock {BERT: Pre-training of Deep Bidirectional Transformers for Language
  Understanding}.
\newblock In {\em Proceedings of NAACL}, pages 4171--4186.

\bibitem[\protect\citename{Ding \bgroup et al.\egroup }2000]{Ding-2000}
Junyan Ding, Luis Gravano, and Narayanan Shivakumar.
\newblock 2000.
\newblock Computing geographical scopes of web resources.
\newblock In {\em Proceedings of VLDV}.

\bibitem[\protect\citename{Doyle}2014]{Doyle-EACL-2014}
Gabriel Doyle.
\newblock 2014.
\newblock Mapping dialectal variation by querying social media.
\newblock In {\em Proceedings of EACL 2014}, pages 98--106.

\bibitem[\protect\citename{Drucker \bgroup et al.\egroup
  }1997]{Drucker-NIPS-1997}
Harris Drucker, Christopher~J.C. Burges, Linda Kaufman, Alex~J. Smola, and
  Vladimir Vapnik.
\newblock 1997.
\newblock Support vector regression machines.
\newblock In {\em Proceedings of NIPS}, pages 155--161.

\bibitem[\protect\citename{Eisenstein \bgroup et al.\egroup
  }2010]{Eisenstein-EMNLP-2010}
Jacob Eisenstein, Brendan O’Connor, Noah~A Smith, and Eric Xing.
\newblock 2010.
\newblock A latent variable model for geographic lexical variation.
\newblock In {\em Proceedings of EMNLP 2010}, pages 1277--1287.

\bibitem[\protect\citename{Friedman}2001]{Friedman-2001}
Jerome~H Friedman.
\newblock 2001.
\newblock Greedy function approximation: a gradient boosting machine.
\newblock {\em Annals of statistics}, pages 1189--1232.

\bibitem[\protect\citename{Fukushima}1980]{Fukushima-BC-1980}
Kunihiko Fukushima.
\newblock 1980.
\newblock Neocognitron: A self-organizing neural network model for a mechanism
  of pattern recognition unaffected by shift in position.
\newblock {\em Biological cybernetics}, 36(4):193--202.

\bibitem[\protect\citename{Gasthaus \bgroup et al.\egroup
  }2010]{Gasthaus-DCC-2010}
Jan Gasthaus, Frank Wood, and Yee~Whye Teh.
\newblock 2010.
\newblock {Lossless Compression Based on the Sequence Memoizer}.
\newblock In {\em Proceedings of DCC}, page 337–345.

\bibitem[\protect\citename{Gelernter and Mushegian}2011]{Gelernter-2011}
Judith Gelernter and Nikolai Mushegian.
\newblock 2011.
\newblock Geo-parsing messages from microtext.
\newblock {\em Transactions in GIS}, 15(6):753--773.

\bibitem[\protect\citename{Georgescu \bgroup et al.\egroup
  }2020]{Georgescu-ArXiv-2020}
Mariana-Iuliana Georgescu, Radu~Tudor Ionescu, Nicolae-Catalin Ristea, and Nicu
  Sebe.
\newblock 2020.
\newblock {Non-linear Neurons with Human-like Apical Dendrite Activations}.
\newblock {\em arXiv preprint arXiv:2003.03229}.

\bibitem[\protect\citename{Gers \bgroup et al.\egroup }2000]{Gers-NC-2000}
Felix~A. Gers, J{\"u}rgen Schmidhuber, and Fred Cummins.
\newblock 2000.
\newblock {Learning to forget: Continual prediction with LSTM}.
\newblock {\em Neural Computation}, 12(10):2451--2471.

\bibitem[\protect\citename{Gim\'{e}nez-P\'{e}rez \bgroup et al.\egroup
  }2017]{Gimenez-EACL-2017}
Rosa~M. Gim\'{e}nez-P\'{e}rez, Marc Franco-Salvador, and Paolo Rosso.
\newblock 2017.
\newblock {Single and Cross-domain Polarity Classification using String
  Kernels}.
\newblock In {\em Proceedings of EACL}, pages 558--563.

\bibitem[\protect\citename{G\u{a}man and Ionescu}2020]{Gaman-ArXiv-2020}
Mihaela G\u{a}man and Radu~Tudor Ionescu.
\newblock 2020.
\newblock {The Unreasonable Effectiveness of Machine Learning in Moldavian
  versus Romanian Dialect Identification}.
\newblock {\em journal={arXiv preprint arXiv:2007.15700}}.

\bibitem[\protect\citename{G\u{a}man \bgroup et al.\egroup
  }2020]{Gaman-VarDial-2020}
Mihaela G\u{a}man, Dirk Hovy, Radu~Tudor Ionescu, Heidi Jauhiainen, Tommi
  Jauhiainen, Krister Lind{\'e}n, Nikola Ljube\v{s}i\'{c}, Niko Partanen,
  Christoph Purschke, Yves Scherrer, and Marcos Zampieri.
\newblock 2020.
\newblock {A Report on the VarDial Evaluation Campaign 2020}.
\newblock In {\em Proceedings of VarDial}.

\bibitem[\protect\citename{Han \bgroup et al.\egroup }2014]{Han-AIR-2014}
Bo~Han, Paul Cook, and Timothy Baldwin.
\newblock 2014.
\newblock Text-based twitter user geolocation prediction.
\newblock {\em Journal of Artificial Intelligence Research}, 49:451--500.

\bibitem[\protect\citename{He \bgroup et al.\egroup }2014]{He-DMOA-2014}
Xinran He, Junfeng Pan, Ou~Jin, Tianbing Xu, Bo~Liu, Tao Xu, Yanxin Shi,
  Antoine Atallah, Ralf Herbrich, Stuart Bowers, et~al.
\newblock 2014.
\newblock Practical lessons from predicting clicks on ads at facebook.
\newblock In {\em Proceedings of ADKDD}, pages 1--9.

\bibitem[\protect\citename{Hochreiter and Schmidhuber}1997]{Hochreiter-NC-1997}
Sepp Hochreiter and J{\"u}rgen Schmidhuber.
\newblock 1997.
\newblock {Long Short-Term Memory}.
\newblock {\em Neural Computation}, 9(8):1735--1780.

\bibitem[\protect\citename{Hochreiter \bgroup et al.\egroup
  }2001]{Hochreiter-FGDRNN-2001}
Sepp Hochreiter, Yoshua Bengio, Paolo Frasconi, J{\"u}rgen Schmidhuber, et~al.
\newblock 2001.
\newblock Gradient flow in recurrent nets: the difficulty of learning long-term
  dependencies.
\newblock {\em A Field Guide to Dynamical Recurrent Neural Networks}, pages
  237--244.

\bibitem[\protect\citename{Hong \bgroup et al.\egroup }2012]{Hong-WWW-2012}
Liangjie Hong, Amr Ahmed, Siva Gurumurthy, Alex~J. Smola, and Kostas
  Tsioutsiouliklis.
\newblock 2012.
\newblock Discovering geographical topics in the twitter stream.
\newblock In {\em Proceedings of WWW}, pages 769--778.

\bibitem[\protect\citename{Hovy and Purschke}2018]{Hovy-EMNLP-2018}
Dirk Hovy and Christoph Purschke.
\newblock 2018.
\newblock {Capturing Regional Variation with Distributed Place Representations
  and Geographic Retrofitting}.
\newblock In {\em Proceedings of EMNLP}, pages 4383--4394.

\bibitem[\protect\citename{Huang \bgroup et al.\egroup }2016]{Huang-CEUS-2016}
Yuan Huang, Diansheng Guo, Alice Kasakoff, and Jack Grieve.
\newblock 2016.
\newblock Understanding us regional linguistic variation with twitter data
  analysis.
\newblock {\em Computers, Environment and Urban Systems}, 59:244--255.

\bibitem[\protect\citename{Ionescu and Butnaru}2017]{Ionescu-VarDial-2017}
Radu~Tudor Ionescu and Andrei~M. Butnaru.
\newblock 2017.
\newblock {Learning to Identify Arabic and German Dialects using Multiple
  Kernels}.
\newblock In {\em Proceedings of VarDial}, pages 200--209.

\bibitem[\protect\citename{Ionescu and Butnaru}2018]{Ionescu-EMNLP-2018}
Radu~Tudor Ionescu and Andrei~M. Butnaru.
\newblock 2018.
\newblock {Improving the results of string kernels in sentiment analysis and
  Arabic dialect identification by adapting them to your test set}.
\newblock In {\em Proceedings of EMNLP}, pages 1084--1090.

\bibitem[\protect\citename{Ionescu and Popescu}2016]{Ionescu-VarDial-2016}
Radu~Tudor Ionescu and Marius Popescu.
\newblock 2016.
\newblock {UnibucKernel: An Approach for Arabic Dialect Identification based on
  Multiple String Kernels}.
\newblock In {\em Proceedings of VarDial}, pages 135--144.

\bibitem[\protect\citename{Ionescu and Popescu}2017]{Ionescu-BEA-2017}
Radu~Tudor Ionescu and Marius Popescu.
\newblock 2017.
\newblock Can string kernels pass the test of time in native language
  identification?
\newblock In {\em Proceedings of BEA-12}, pages 224--234.

\bibitem[\protect\citename{Ionescu \bgroup et al.\egroup
  }2014]{Ionescu-EMNLP-2014}
Radu~Tudor Ionescu, Marius Popescu, and Aoife Cahill.
\newblock 2014.
\newblock {Can characters reveal your native language? A language-independent
  approach to native language identification}.
\newblock In {\em Proceedings of EMNLP}, pages 1363--1373.

\bibitem[\protect\citename{Ionescu \bgroup et al.\egroup
  }2016]{Ionescu-COLI-2016}
Radu~Tudor Ionescu, Marius Popescu, and Aoife Cahill.
\newblock 2016.
\newblock String kernels for native language identification: Insights from
  behind the curtains.
\newblock {\em Computational Linguistics}, 42(3):491--525.

\bibitem[\protect\citename{Jones}2015]{Jones-AS-2015}
Taylor Jones.
\newblock 2015.
\newblock {Toward a description of African American Vernacular English dialect
  regions using ``Black Twitter''}.
\newblock {\em American Speech}, 90(4):403--440.

\bibitem[\protect\citename{Kim \bgroup et al.\egroup }2016]{Kim-AAAI-2016}
Yoon Kim, Yacine Jernite, David Sontag, and Alexander~M. Rush.
\newblock 2016.
\newblock {Character-Aware Neural Language Models}.
\newblock In {\em Proceedings of AAAI}, pages 2741--2749.

\bibitem[\protect\citename{Kim}2014]{Kim-EMNLP-2014}
Yoon Kim.
\newblock 2014.
\newblock {Convolutional Neural Networks for Sentence Classification}.
\newblock In {\em Proceedings of EMNLP}, pages 1746--1751.

\bibitem[\protect\citename{Kingma and Ba}2015]{Kingma-ICLR-2014}
Diederik~P. Kingma and Jimmy Ba.
\newblock 2015.
\newblock Adam: A method for stochastic optimization.
\newblock In {\em Proceedings of ICLR}.

\bibitem[\protect\citename{Kinsella \bgroup et al.\egroup }2011]{Kinsella-2011}
Sheila Kinsella, Vanessa Murdock, and Neil O'Hare.
\newblock 2011.
\newblock {``I'm eating a sandwich in Glasgow'' modeling locations with
  tweets}.
\newblock In {\em Proceedings of SMUC}, pages 61--68.

\bibitem[\protect\citename{Krizhevsky \bgroup et al.\egroup
  }2012]{Krizhevsky-NIPS-2012}
Alex Krizhevsky, Ilya Sutskever, and Geoffrey~E. Hinton.
\newblock 2012.
\newblock {ImageNet Classification with Deep Convolutional Neural Networks}.
\newblock In {\em Proceddings of NIPS}, pages 1097--1105.

\bibitem[\protect\citename{LeCun \bgroup et al.\egroup }1989]{LeCun-NC-1989}
Yann LeCun, Bernhard Boser, John~S. Denker, Donnie Henderson, Richard~E.
  Howard, Wayne Hubbard, and Lawrence~D. Jackel.
\newblock 1989.
\newblock {Backpropagation Applied to Handwritten Zip Code Recognition}.
\newblock {\em Neural Computation}, 1(4):541--551.

\bibitem[\protect\citename{LeCun \bgroup et al.\egroup }2004]{LeCun-CVPR-2004}
Yann LeCun, Fu~Jie Huang, and Leon Bottou.
\newblock 2004.
\newblock Learning methods for generic object recognition with invariance to
  pose and lighting.
\newblock In {\em Proceedings of CVPR}, volume~2, pages II--104.

\bibitem[\protect\citename{Li}2010]{Li-UAI-2010}
Ping Li.
\newblock 2010.
\newblock {Robust Logitboost and Adaptive Base Class (ABC) Logitboost}.
\newblock In {\em Proceedings of UAI}, pages 302--311.

\bibitem[\protect\citename{Lieberman \bgroup et al.\egroup
  }2010]{Lieberman-ICDE-2010}
Michael~D Lieberman, Hanan Samet, and Jagan Sankaranarayanan.
\newblock 2010.
\newblock Geotagging with local lexicons to build indexes for
  textually-specified spatial data.
\newblock In {\em In Proceedings of ICDE}, pages 201--212.

\bibitem[\protect\citename{Ljube{\v{s}}i{\'c} \bgroup et al.\egroup
  }2016]{Ljubesic-COLING-2016}
Nikola Ljube{\v{s}}i{\'c}, Tanja Samard{\v{z}}i{\'c}, and Curdin Derungs.
\newblock 2016.
\newblock {{T}weet{G}eo - A Tool for Collecting, Processing and Analysing
  Geo-encoded Linguistic Data}.
\newblock In {\em Proceedings of COLING}, pages 3412--3421.

\bibitem[\protect\citename{Lodhi \bgroup et al.\egroup }2001]{Lodhi-NIPS-2001}
Huma Lodhi, John Shawe-Taylor, Nello Cristianini, and Christopher~J.C.H.
  Watkins.
\newblock 2001.
\newblock {Text Classification Using String Kernels}.
\newblock In {\em Proceedings of NIPS}, pages 563--569.

\bibitem[\protect\citename{Masala \bgroup et al.\egroup }2017]{Masala-KES-2017}
Mihai Masala, Stefan Ruseti, and Traian Rebedea.
\newblock 2017.
\newblock Sentence selection with neural networks using string kernels.
\newblock In {\em Proceedings of KES}, pages 1774--1782.

\bibitem[\protect\citename{Onose \bgroup et al.\egroup
  }2019]{Onose-VarDial-2019}
Cristian Onose, Dumitru-Clementin Cercel, and \c{S}tefan Tr\u{a}u\c{s}an-Matu.
\newblock 2019.
\newblock {SC-UPB at the VarDial 2019 Evaluation Campaign: Moldavian vs.
  Romanian Cross-Dialect Topic Identification}.
\newblock In {\em Proceedings of VarDial}, pages 172--177.

\bibitem[\protect\citename{Popescu and Ionescu}2013]{Popescu-BEA8-2013}
Marius Popescu and Radu~Tudor Ionescu.
\newblock 2013.
\newblock {The Story of the Characters, the DNA and the Native Language}.
\newblock In {\em Proceedings of BEA-8}, pages 270--278.

\bibitem[\protect\citename{Popescu \bgroup et al.\egroup
  }2017]{Popescu-KES-2017}
Marius Popescu, Cristian Grozea, and Radu~Tudor Ionescu.
\newblock 2017.
\newblock {HASKER: An efficient algorithm for string kernels. Application to
  polarity classification in various languages}.
\newblock In {\em Proceedings of KES}, pages 1755--1763.

\bibitem[\protect\citename{Qin \bgroup et al.\egroup
  }2010]{Qin-SIGSPATIAL-2010}
Teng Qin, Rong Xiao, Lei Fang, Xing Xie, and Lei Zhang.
\newblock 2010.
\newblock An efficient location extraction algorithm by leveraging web
  contextual information.
\newblock In {\em Proceedings of GIS}, pages 53--60.

\bibitem[\protect\citename{Quercini \bgroup et al.\egroup
  }2010]{Quercini-SIGSPATIAL-2010}
Gianluca Quercini, Hanan Samet, Jagan Sankaranarayanan, and Michael~D
  Lieberman.
\newblock 2010.
\newblock Determining the spatial reader scopes of news sources using local
  lexicons.
\newblock In {\em Proceedings of GIS}, pages 43--52.

\bibitem[\protect\citename{Radford \bgroup et al.\egroup
  }2018]{Radford-Arxiv-2018}
Alec Radford, Karthik Narasimhan, Tim Salimans, and Ilya Sutskever.
\newblock 2018.
\newblock Improving language understanding with unsupervised learning.
\newblock {\em Technical report, OpenAI}.

\bibitem[\protect\citename{Rahimi \bgroup et al.\egroup
  }2017]{Rahimi-Arxiv-2017}
Afshin Rahimi, Trevor Cohn, and Timothy Baldwin.
\newblock 2017.
\newblock A neural model for user geolocation and lexical dialectology.
\newblock {\em arXiv preprint arXiv:1704.04008}.

\bibitem[\protect\citename{Roller \bgroup et al.\egroup
  }2012]{Roller-EMNLP-2012}
Stephen Roller, Michael Speriosu, Sarat Rallapalli, Benjamin Wing, and Jason
  Baldridge.
\newblock 2012.
\newblock Supervised text-based geolocation using language models on an
  adaptive grid.
\newblock In {\em Proceedings of EMNLP}, pages 1500--1510.

\bibitem[\protect\citename{Rout \bgroup et al.\egroup }2013]{Rout-ACM-2013}
Dominic Rout, Kalina Bontcheva, Daniel Preo{\c{t}}iuc-Pietro, and Trevor Cohn.
\newblock 2013.
\newblock {Where's@ wally? A Classification Approach to Geolocating Users Based
  on their Social Ties}.
\newblock In {\em Proceedings of HT}, pages 11--20.

\bibitem[\protect\citename{Smola and Sch{\"o}lkopf}2004]{Smola-SC-2004}
Alex~J. Smola and Bernhard Sch{\"o}lkopf.
\newblock 2004.
\newblock A tutorial on support vector regression.
\newblock {\em Statistics and computing}, 14(3):199--222.

\bibitem[\protect\citename{Sutskever \bgroup et al.\egroup
  }2011]{Sutskever-ICML-2011}
Ilya Sutskever, James Martens, and Geoffrey Hinton.
\newblock 2011.
\newblock {Generating Text with Recurrent Neural Networks}.
\newblock In {\em Proceedings of ICML}, pages 1017--1024.

\bibitem[\protect\citename{Szmrecsanyi}2008]{Szmrecsanyi-IJHAC-2008}
Benedikt Szmrecsanyi.
\newblock 2008.
\newblock Corpus-based dialectometry: Aggregate morphosyntactic variability in
  british english dialects.
\newblock {\em International Journal of Humanities and Arts Computing},
  2(1-2):279--296.

\bibitem[\protect\citename{Tudoreanu}2019]{Tudoreanu-VarDial-2019}
Diana Tudoreanu.
\newblock 2019.
\newblock {DTeam @ VarDial 2019: Ensemble based on skip-gram and triplet loss
  neural networks for Moldavian vs. Romanian cross-dialect topic
  identification}.
\newblock In {\em Proceedings of VarDial}, pages 202--208.

\bibitem[\protect\citename{Vaswani \bgroup et al.\egroup
  }2017]{Vaswani-NIPS-2017}
Ashish Vaswani, Noam Shazeer, Niki Parmar, Jakob Uszkoreit, Llion Jones,
  Aidan~N Gomez, {\L}ukasz Kaiser, and Illia Polosukhin.
\newblock 2017.
\newblock Attention is all you need.
\newblock In {\em Proceedings of NIPS}, pages 5998--6008.

\bibitem[\protect\citename{Weiss \bgroup et al.\egroup }2018]{Weiss-ACL-2018}
Gail Weiss, Yoav Goldberg, and Eran Yahav.
\newblock 2018.
\newblock {On the Practical Computational Power of Finite Precision RNNs for
  Language Recognition}.
\newblock In {\em Proceedings of ACL}, pages 740--745.

\bibitem[\protect\citename{Werbos}1988]{Werbos-NN-1988}
Paul~J. Werbos.
\newblock 1988.
\newblock Generalization of backpropagation with application to a recurrent gas
  market model.
\newblock {\em Neural Networks}, 1(4):339--356.

\bibitem[\protect\citename{Wieling \bgroup et al.\egroup
  }2011]{Wieling-PLoS-2011}
Martijn Wieling, John Nerbonne, and R.~Harald Baayen.
\newblock 2011.
\newblock Quantitative social dialectology: Explaining linguistic variation
  geographically and socially.
\newblock {\em PloS One}, 6(9):e23613.

\bibitem[\protect\citename{Wing and Baldridge}2011]{Wing-HLT-2011}
Benjamin Wing and Jason Baldridge.
\newblock 2011.
\newblock Simple supervised document geolocation with geodesic grids.
\newblock In {\em Proceedings of ACL}, pages 955--964.

\bibitem[\protect\citename{Wood \bgroup et al.\egroup }2009]{Wood-ICML-2009}
Frank Wood, C\'{e}dric Archambeau, Jan Gasthaus, Lancelot James, and Yee~Whye
  Teh.
\newblock 2009.
\newblock {A Stochastic Memoizer for Sequence Data}.
\newblock In {\em Proceedings of ICML}, pages 1129--1136.

\bibitem[\protect\citename{Zhang \bgroup et al.\egroup }2015]{Zhang-NIPS-2015}
Xiang Zhang, Junbo Zhao, and Yann LeCun.
\newblock 2015.
\newblock {Character-level Convolutional Networks for Text Classification}.
\newblock In {\em Proceedings of NIPS}, pages 649--657.

\end{thebibliography}

\end{document}